\documentclass[10pt,twocolumn,letterpaper]{article}

\usepackage{cvpr}
\usepackage{times}
\usepackage{epsfig}
\usepackage{graphicx}
\usepackage{amsmath}
\usepackage{amssymb}

\usepackage{booktabs}

\usepackage{algorithm}
\usepackage{algpseudocode}

\usepackage{multirow}

\usepackage[pagebackref=true,breaklinks=true,letterpaper=true,colorlinks,bookmarks=false]{hyperref}

\cvprfinalcopy 

\def\cvprPaperID{6637} 

\ifcvprfinal\fi
\begin{document}

\title{ Weakly Supervised Attention Pyramid Convolutional Neural Network \\ for Fine-Grained Visual Classification}

\author{Yifeng Ding\textsuperscript{\rm 1} \quad Shaoguo Wen\textsuperscript{\rm 1} \quad Jiyang Xie\textsuperscript{\rm 1} \quad Dongliang Chang\textsuperscript{\rm 1}\\
Zhanyu Ma\textsuperscript{\rm 1} \quad Zhongwei Si\textsuperscript{\rm 1} \quad Haibin Ling\textsuperscript{\rm 2}\\
{\tt\small \textsuperscript{\rm 1}Beijing University of Posts and Telecommunications, China} \quad
{\tt\small \textsuperscript{\rm 2}Stony Brook University, NewYork}\\
}

\maketitle

\begin{abstract}
Classifying the sub-categories of an object from the same super-category (\eg, bird species, car and aircraft models) in fine-grained visual classification (FGVC) highly relies on discriminative feature representation and accurate region localization. Existing approaches mainly focus on distilling information from high-level features. In this paper, however, we show that by integrating low-level information (\eg, color, edge junctions, texture patterns), performance can be improved with enhanced feature representation and accurately located discriminative regions. Our solution, named Attention Pyramid Convolutional Neural Network (AP-CNN), consists of a) a pyramidal hierarchy structure with a top-down feature pathway and a bottom-up attention pathway, and hence learns both high-level semantic and low-level detailed feature representation, and b) an ROI guided refinement strategy with ROI guided dropblock and ROI guided zoom-in, which refines features with discriminative local regions enhanced and background noises eliminated. The proposed AP-CNN can be trained end-to-end, without the need of additional bounding box/part annotations. Extensive experiments on three commonly used FGVC datasets (CUB-200-2011, Stanford Cars, and FGVC-Aircraft) demonstrate that our approach can achieve state-of-the-art performance. Code available at \url{http://dwz1.cc/ci8so8a}
\end{abstract}

\section{Introduction}\label{Introduction}
The fine-grained visual classification (FGVC) task focuses on differentiating sub-categories of the objects from the same super-category (\eg, bird species, cars and aircrafts models). It has attracted extensive attention recently due to a wide range of applications such as expert-level image recognition~\cite{xiao2015application}, rich image captioning~\cite{johnson2016densecap}, intelligent retail~\cite{baz2016context}, and intelligent transportation~\cite{yang2015large}. Different from the traditional image classification task, images from different sub-classes in the FGVC problems share close similarities. At the same time, it differs from the face recognition task, because the faces are aligned into similar directions in face recognition while different poses are often occurred in FGVC. As a result, the challenging and distinctive keystones of the FGVC problem are a) high intra-class variance: objects that belong to the same category usually present significantly different poses and viewpoints; and b) low inter-class variance: the visual differences among the subordinate classes are often subtle as they belong to the same super-category. 

\begin{figure}[!t]
\begin{center}
   \includegraphics[width=0.95\linewidth]{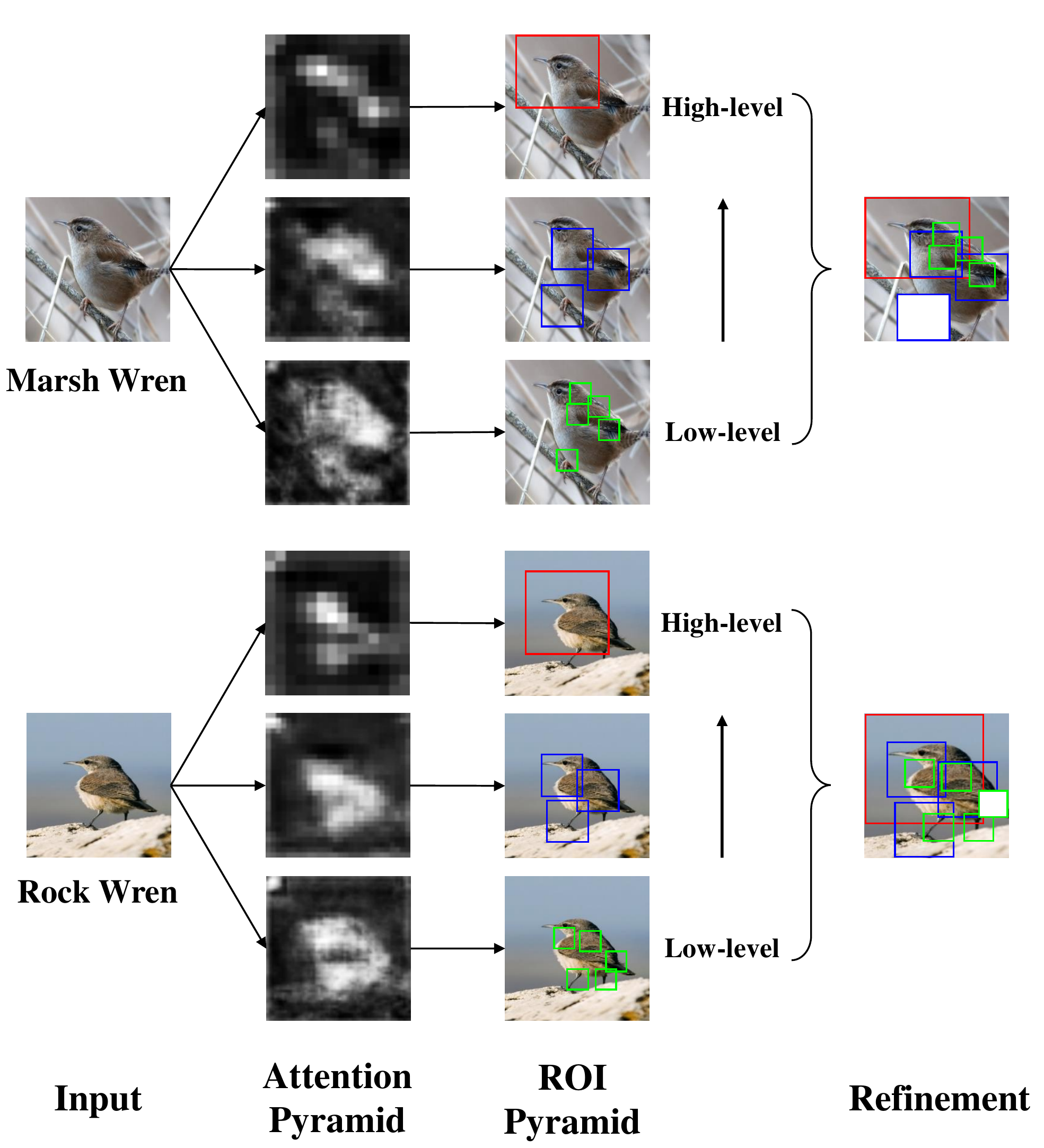}
\end{center}
   \caption{The discriminative regions of interest from different pyramidal hierarchy learned by AP-CNN for two bird species of ``wren''. It can be observed that the low-level information can capture more subtle parts to distinguish the birds, \eg, the texture of wings and the shape of claws. Refinement is conducted on features with background noises eliminated and discriminative parts enhanced. }
\label{fig:fig1}
\end{figure}

In order to address the aforementioned challenges, early solutions~\cite{chai2013symbiotic, zhang2014part, huang2016part} introduce additional bounding box/part annotations to help locate the target object and align the components in an image. Although effective, these methods are not optimal for FGVC as human annotations can be very time consuming and need professional knowledge. 
Recent methods solve this problem in a weakly supervised manner. These approaches can be classified into two categories: a) feature encoding methods~\cite{lin2015bilinear, kong2017low, yu2018hierarchical} that extract the fine-grained features by encoding a highly parameterized representation of the features, and b) region locating methods~\cite{zheng2017learning, fu2017look, yang2018learning, ge2019weakly} that figure out the discriminative regions by learning part detectors, and then conduct refinement such as cropping and amplifying the attended parts on multiple stages. 

Although promising results have been reported in the above studies, further improvement suffers from lack of using low-level information. Our study shows that low-level information (\eg, color, edge junctions, texture patterns~\cite{zeiler2014visualizing}), is indeed essential in the FGVC task. Specifically, as the structure of CNN getting deeper, the neurons in high layers are strongly respond to entire images and rich in semantics, but inevitably lose detailed information from small discriminative regions. Figure~\ref{fig:fig1} shows examples of differences in activations extracted from diverse layers. such detailed information, \emph{i.e.} the low-level information, is helpful in the FGVC task as it reflects the subtle difference within various sub-classes, and is invariable no matter how the pose or viewpoint changes. Existing FGVC methods pay much attention to high-level features, and in this work, we use additional enhanced low-level information as supplement.

The motivation of this paper is to effectively integrate both the high-level semantic and the low-level detailed information for fine-grained classification. To this end, we propose a novel attention pyramid convolutional neural network (AP-CNN), which jointly learns multi-level information and refined feature representations without using bounding box/part annotations. The proposed method can accurately locate the discriminative local regions as well as reduce the background noise.
The main contribution can be summarized as follows:

1) We propose a novel attention pyramid convolutional neural network (AP-CNN) by building an enhanced pyramidal hierarchy, which combines a top-down pathway of features and a bottom-up pathway of attentions, and thus learns both high-level semantic and low-level detailed feature representations.

2) We propose ROI guided refinement consisting of ROI guided dropblock and ROI guided zoom-in to further refine the features. The dropblock operation helps to locate more discriminative local regions, and the zoom-in operation aligns features with background noises eliminated. 

3) We conduct extensive experiments on three commonly used FGVC datasets (CUB-200-2011~\cite{wah2011caltech}, Stanford-Cars~\cite{krause20133d}, and FGVC-Aircraft~\cite{maji2013fine}). Visualization and ablative studies are further conducted to draw insights into our method. The results demonstrate that our model can significantly improve the accuracy of fine-grained classification.

\begin{figure*}[!t]
\begin{center}
   \includegraphics[width=1.0\linewidth]{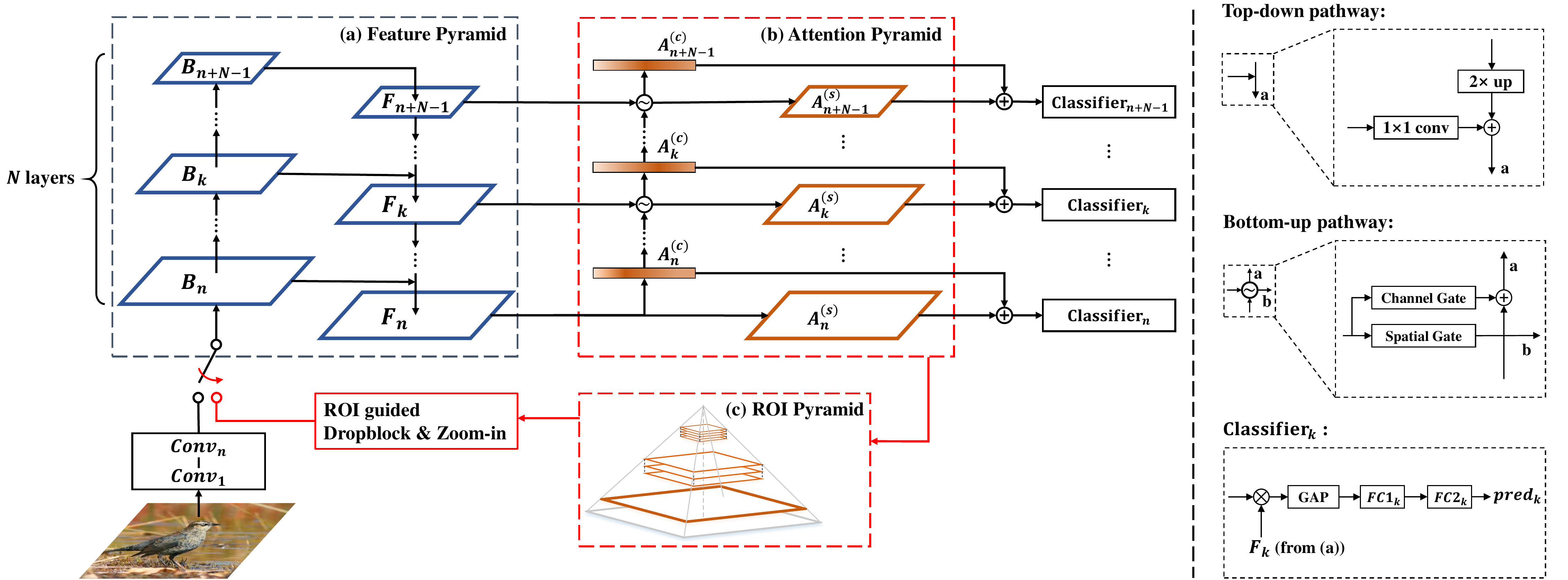}
\end{center}
   \caption{Illustration of AP-CNN. (a) FPN backbone. (b)Attention pyramid module. (c) ROI pyramid. The refinement operation is shown in the red flow and conducted on the low-level features $B_n$. In this figure, the feature maps are indicated by blue outlines and the channel/spatial attentions are indicated in orange. The structure details of the top-down and the bottom-up pathways, and the classifiers are illustrated on the right. $\oplus$ represents broadcasting addition and $\otimes$ represents element-wise multiplication.}
\label{fig:overview}
\end{figure*}

\section{Related Work}\label{RelatedWork}
\textbf{Methods Using Multi-level Features.} Features from different layers are commonly used in detection and segmentation tasks. NoCs~\cite{ren2016object} extracts feature maps on input images with different scales and then conduct feature fusion. FCN~\cite{long2015fully}, U-Net~\cite{ronneberger2015u}, FPN~\cite{lin2017feature} fuse information from lower layers to high-level features through skip-connections. Meanwhile, HyperNet~\cite{kong2016hypernet}, ParseNet~\cite{liu2015parsenet}, and ION~\cite{bell2016inside} concatenate features of multiple layers before computing predictions.
SSD~\cite{liu2016ssd} and MS-CNN~\cite{cai2016unified} predict individual target locations at multiple layers without combining features. 

In this paper, we use the multi-level features in the FGVC task for better classification and weakly supervised detection. Besides, we further enhance the multi-level features by establishing strong correlations between them. This is done through a top-down feature pathway delivering the semantic information from high levels to low levels, combined with a bottom-up attention pathway carrying low-level information back to the top.

\textbf{Weakly Supervised Fine-grained Classification.} We define weakly supervised learning in FGVC task as methods without bounding box/part annotations. This setting is generally applied in recent methods due to its feasibility in real-world scenarios.

Feature encoding-based approaches~\cite{lin2015bilinear, gao2016compact, kong2017low, yu2018hierarchical} encode higher order information on features. The classical benchmark, Bilinear-CNN~\cite{lin2015bilinear}, utilizes a bilinear pooling operation to aggregate the pairwise interactions between features in two independent CNNs, and computes the outer product over the output feature channels of the two streams to capture the second-order information. Further works~\cite{gao2016compact, kong2017low} reduce the huge computation cost of the outer product by a single-stream output and adopting low-rank approximation to the covariance matrix, respectively. Moreover, \cite{yu2018hierarchical} proposes a cross-layer bilinear pooling approach to capture the inter-layer part feature relations.

Region locating methods~\cite{fu2017look, zheng2017learning, yang2018learning, ge2019weakly}, another common way in FGVC, generally apply localization networks to detect discriminative regions in images. RA-CNN \cite{fu2017look} is proposed to zoom in discriminative local regions learned by a novel attention proposal network. Meanwhile, MA-CNN \cite{zheng2017learning} generates multiple object parts by clustering channels of feature maps into different groups. NTS \cite{yang2018learning} enables a navigator agent as the region proposal network to detect multiple most informative regions under the guidance from a teacher agent. WS-CPM \cite{ge2019weakly} develops a novel pipeline combined with object detection and segmentation using Mask R-CNN, followed by a bi-directional long short-term memory (LSTM) network to integrate and encode the partial information.

The most relevant work to ours comes from NTS ~\cite{yang2018learning}, which also applies pyramidal features to the FGVC task. However, the NTS network only learns the region localizers by simply applying the FPN~\cite{lin2017feature} structure on the CNN, which poses challenges to accurate region localization in the weakly supervised way. Besides, it ignores the fact that these pyramidal features can also contribute to the classification. 

Compared with NTS, the advantages of our work are two-folds. First, we introduce the pyramid hierarchy to the FGVC task and further enhance its representation. We use these multi-level information not only for precise region localization but also for better classification. Second, we conduct refinement that takes full advantage of multi-level features, by using small ROIs learned from low-level features for dropblock operation, and using bounding rectangle merged by ROIs from all levels for zoom-in operation.



\section{Attention Pyramid Convolutional Neural Network}\label{Approach}
In this section, we introduce the proposed Attention Pyramid Convolutional Neural Network (AP-CNN) for fine-grained classification. AP-CNN is a two-stage network that respectively takes coarse full images and refined features as input. These two stages, which we define as the raw-stage and the refined-stage, share the same network architecture with the same parameters to extract information from both the coarse and the refined inputs. 

An overview of the proposed AP-CNN is illustrated in Figure~\ref{fig:overview}. Feature and attention pyramid structure, and ROI guided refinement are conducted for improving performance. First, the feature and attention pyramid structure takes coarse images as input, which generates the pyramidal features and the pyramidal attentions by establishing hierarchy on the basic CNN following a top-down feature pathway and a bottom-up attention pathway. Second, once the spatial attention pyramid has been obtained from the raw input, the region proposal network (RPN) proceeds to generate the pyramidal regions of interest (ROIs) in a weakly supervised way. Then the ROI guided refinement is conducted on low-level features with a) the ROI guided dropblock which erases the most discriminative regions selected from small-scaled ROIs, and b) the ROI guided zoom-in which locates the major regions merged from all ROIs. Third, the refined features are sent into the refined-stage to distill more discriminative information. Both stages set individual classifiers for each pyramid level, and the final classification result is averaged over the raw-stage predictions and the refined-stage predictions. 
Note that the AP-CNN can be trained end-to-end, and the framework is flexible in the CNN backbone structure (\eg, AlexNet~\cite{krizhevsky2012imagenet}, VGG~\cite{simonyan2014very} and ResNet~\cite{he2016deep}). In this paper, we present results using VGG16 and ResNet50.

\begin{figure*}[!t]
\begin{center}
   \includegraphics[width=0.9\linewidth]{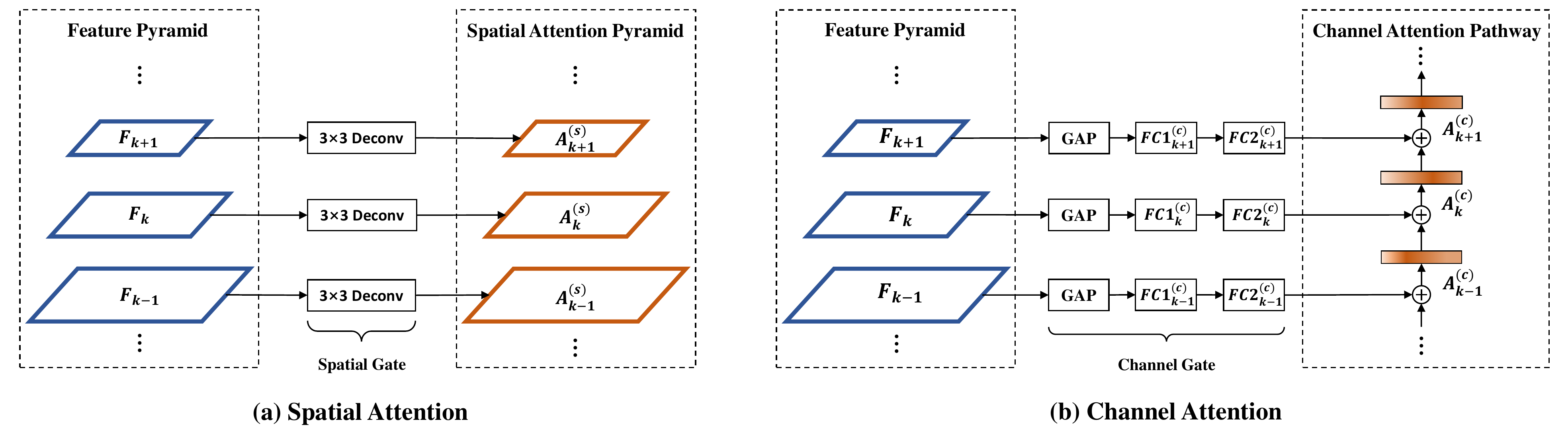}
\end{center}
   \caption{Process of getting (a) spatial attention pyramid and (b) channel attention pathway. }
\label{fig:fig23}
\end{figure*}

\subsection{Attention Pyramid Model}\label{apm}
\textbf{Motivation.}
Our goal is to leverage both the semantic and the detailed information for improving FGVC performance. Specifically, CNN backbones apply a series of convolution blocks, we denote the output feature maps of blocks with the different spatial size as $\{\boldsymbol{B}_1, \boldsymbol{B}_2, \ldots, \boldsymbol{B}_l\}$, where $l$ indicates the number of blocks. Traditional methods use the last feature map $\boldsymbol{B}_l$ for classification. It contains strong semantics but lacks detailed information, which is adverse to the FGVC task. FPN chooses part of these features and generates $N$ corresponding feature hierarchy $\{\boldsymbol{F}_n, \boldsymbol{F}_{n+1}, \ldots, \boldsymbol{F}_{n+N-1}\}$ $(1 \le n \le n+N-1 \le l)$ by applying a) a top-down pathway after $\boldsymbol{B}_{n+N-1}$, which upsamples spatially coarser but semantically stronger feature maps from higher pyramid levels to lower pyramid levels, and b) lateral connections between corresponding features $\boldsymbol{B}_k \rightarrow \boldsymbol{F}_k$ $(k = n,\ n+1,\ \ldots,\ n+N-1)$ to maintain the backbone information. The pyramidal features can locate samples on different scales, which is also beneficial to the FGVC task, to focus on the subtle differences of objects from different scales.

We further enhance the FPN structure by introducing an additional attention hierarchy $\{\boldsymbol{A}_n, \boldsymbol{A}_{n+1}, \ldots, \boldsymbol{A}_{n+N-1}\}$ upon the pyramidal features, which consists of (a) pyramidal spatial attentions $\{\boldsymbol{A}_n^{(s)}, \boldsymbol{A}_{n+1}^{(s)}, \ldots, \boldsymbol{A}_{n+N-1}^{(s)}\}$ to locate discriminative regions from different scales, and (b) pyramidal channel attentions $\{\boldsymbol{A}_n^{(c)}, \boldsymbol{A}_{n+1}^{(c)}, \ldots, \boldsymbol{A}_{n+N-1}^{(c)}\}$ to embed channel correlations and diliver local information from lower pyramid levels to higher pyramid levels in an additional bottom-up pathway.

\textbf{Spatial Gate and Spatial Attention Pyramid.} As shown in Figure~\ref{fig:fig23}(a), each building block takes the corresponding feature map $\boldsymbol{F}_k$ as input and generates a spatial attention mask $\boldsymbol{A}_{k}^{(s)}$. Specifically, each feature map $\boldsymbol{F}_k$ first goes through a $3 \times 3$ deconvolution layer with one output channel to squeeze spatial information. Then each element of the spatial attention mask $\boldsymbol{A}_{k}^{(s)}$ is normalized to the interval (0,1) using the sigmoid function to reflect the spatial importance:
\begin{equation}
   \boldsymbol{A}_{k}^{(s)} = \sigma (\boldsymbol{v}_c \ast \boldsymbol{F}_{k}).
   \label{eq:spatial_1}
\end{equation}
Here $\sigma$ refers to the sigmoid function, while $\ast$ denotes deconvolution and $\boldsymbol{v}_c$ represents convolution kernel. As a result, we get spatial attention pyramid $\{\boldsymbol{A}_{n}^{(s)}, \boldsymbol{A}_{n+1}^{(s)}, \ldots, \boldsymbol{A}_{n+N-1}^{(s)}\}$ based on multi-scale feature maps. We use these spatial activations to generate the ROI pyramid and conduct further refinement on features, which is described as below.

\textbf{Channel Gate and Channel Attention Pathway.} Inspired by SE-Net \cite{hu2018squeeze}, channel attentions $\{\boldsymbol{A}_{n}^{(c)}, \boldsymbol{A}_{n+1}^{(c)},  \ldots,$ $ \boldsymbol{A}_{n+N-1}^{(c)}\}$ are gained from corresponding feature maps in the feature pyramid by operating the global average pooling (GAP) combined with two fully-connected (FC) layers. The channel attention mask can be represented as:
\begin{equation}
   \boldsymbol{A}_{k}^{(c)} = \sigma (\boldsymbol{W_2} \cdot {\rm ReLU} (\boldsymbol{W_1} \cdot {\rm GAP}(\boldsymbol{F}_{k}))),
   \label{eq:channel_1}
\end{equation}
where ${\rm GAP}(\cdot)$ is the global average pooling function:
\begin{equation}
   {\rm GAP}(\boldsymbol{F}_{k}) = \frac{1}{W\times H}\sum_{i=1}^{H}\sum_{j=1}^{W}\boldsymbol{F}_{k}(i,j).
   \label{eq:channel_2}
\end{equation}
Here $\sigma$ and $ReLU$ refer to the sigmoid and the ReLU function, respectively. The dot product denotes element-wise multiplication. $W$ and $H$ represent the spatial dimensions of $\boldsymbol{F}_{k}$. $\boldsymbol{W_1}$ and $\boldsymbol{W_2}$ are the weight matrices of two FC layers. 
In our framework, channel attentions play a different role from the spatial attention pyramid as they are settled for delivering low-level detailed information in a bottom-up pathway from lower pyramid levels to higher pyramid levels. Figure~\ref{fig:fig23}(b) shows the flow diagram.

\textbf{Classifier.} We use the learned attentions to weight features $\boldsymbol{F}_{k}$, and get $\boldsymbol{F}_{k}^{'}$ for classification:
\begin{equation}
   \boldsymbol{F}_{k}^{'} = \boldsymbol{F}_{k} \cdot (\boldsymbol{A}_{k}^{(s)} \oplus \boldsymbol{A}_{k}^{(c)}),
   \label{eq:channel_1}
\end{equation}
where $\oplus$ represents the addition operation using broadcasting semantics. Individual classifiers with a GAP layer and two FC layers are settled to make final predictions.

\begin{figure*}[t]
\begin{center}
   \includegraphics[width=0.95\linewidth]{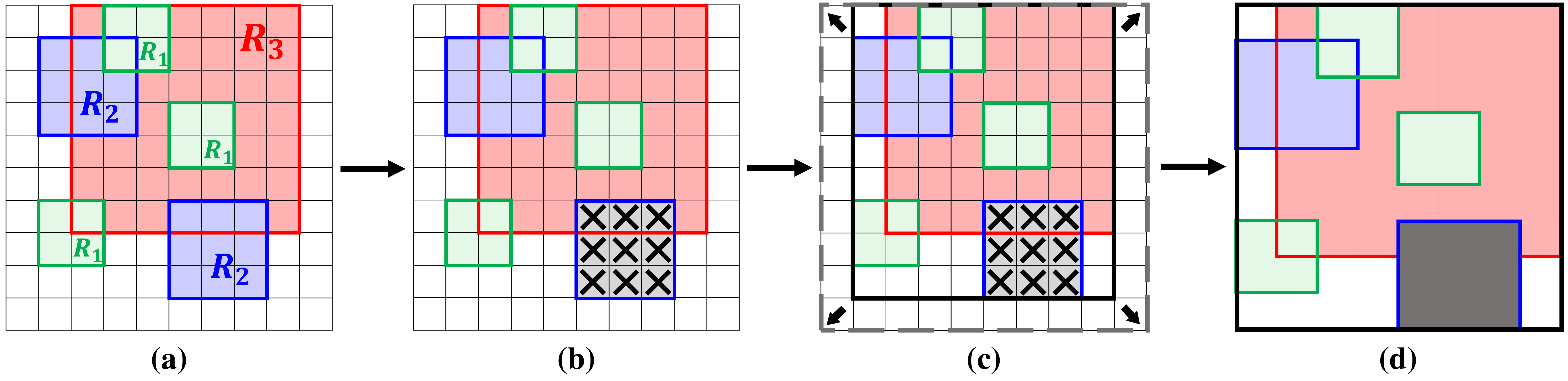}
\end{center}
   \caption{ROI guided refinement based on Algorithm~\ref{alg:refinement}. (a) Low-level feature map $\boldsymbol{B}_n$ with ROI pyramid $R_{all}$ ($R_1, R_2, R_3$ for example). (b) ROI guided Dropblock. (c) ROI guided zoom-in. (d) Refined feature map $\boldsymbol{Z}_n$ with local regions masked and background noises eliminated. Rectangles with different colors are from distinctive levels in the pyramidal hierarchy, while the gray one represents the dropped region.}
\label{fig:fig5}
\end{figure*}

\begin{algorithm}[t]
\caption{ ROI guided refinement algorithm.}  
\label{alg:refinement}  
\begin{algorithmic}[1]
   \Require
      Low-level feature map $\boldsymbol{B}_n$; 
      ROI pyramid $R_{all}=\{R_n, R_{n+1}, \ldots, R_{n+N-1}\}$; 
      Dropblock candidates selection probabilities $P = \{p_n, p_{n+1}, \ldots, p_{n+N-1}\}$;
    \Ensure
      Refined feature map $\boldsymbol{Z}_n$; 
    \If {training}
      \State Randomly select $R_s$ from $R_{all}$ ($n \leq s \leq n+N-1$) with probabilities $P$;
      \State Randomly select $r_s$ from $R_{s} = \{r_{s,1}, r_{s,2}, \ldots,$ $r_{s,\xi_s}\}$ with equal probability;
      \State Calculate dropblock feature $\boldsymbol{D}_n$ according to Eq. \ref{eq:dropblock feature};
    \Else
      \State $\boldsymbol{D}_n$ = $\boldsymbol{B}_n$;
    \EndIf
    \State Merge $R_{all}$ and obtain the bounding box;
    \State Crop the region and enlarge as $\boldsymbol{Z}_n$ by Eq. \ref{eq:crop_resize}; \\
    \Return $\boldsymbol{Z}_n$;
\end{algorithmic}
\end{algorithm}

\subsection{ROI Guided Refinement}\label{refinement}
\textbf{ROI Pyramid.} Region proposal network (RPN) \cite{ren2015faster} is a widely used structure in visual detection to locate possible informative regions. Recent RPN designs~\cite{lin2017feature} are conducted on a single-scale convolutional feature map or multi-scale convolutional feature maps. Anchors of multiple pre-defined scales and aspect ratios are designed to cover objects of different shapes.

In this work, we adapt RPN on the spatial attention pyramid, rather than the multi-scale feature maps. We assign the anchors with a single scale and ratio to each pyramidal level according to its convolutional receptive field, and adopt non-maximum suppression (NMS) on the region proposals to reduce region redundancy. Specifically, for each pyramid level $k$, we select the top-$\xi_k$ informative regions $R_k = \{r_{k,1}, r_{k,2}, \ldots, r_{k,\xi_k}\}$ based on the responding activation values of anchors in the spatial attention mask, and then form the region pyramid $R_{all} = \{R_n, R_{n+1}, \ldots, R_{n+N-1} \}$. As a result, we get an ROI pyramid in the raw-stage, and then conduct ROI guided refinement (ROI guided dropblock and ROI guided zoom-in) on the pyramid bottom features $\boldsymbol{B}_n$ to further improve the performance in the refined-stage. Algorithm~\ref{alg:refinement} shows the main procedure of the refinement operations and Figure~\ref{fig:fig5} illustrates a visualized example.

\textbf{ROI Guided Dropblock.} Overfitting is a common problem in deep-learning, especially in the FGVC task as each species only contains small number of images. \cite{ghiasi2018dropblock} proposes the dropblock strategy by dropping continuous regions randomly on feature maps to remove certain semantic information and consequently enforcing network to learn information on the remaining units. In this paper, we randomly select an ROI union $R_s$ from $R_{all}$ ($n\leq s\leq n+N-1$) with probabilities $P = \{p_n, p_{n+1}, \ldots, p_{n+N-1}\}$. Then we randomly choose an informative region $r_s \in R_s$ with equal probability from the selected $R_s$. We scale $r_s$ into the same sampling rate as $\boldsymbol{B}_n$, and obtain drop mask $M$ by setting activations in the ROI region to zero:
\begin{equation}
   M(i,j)=\left\{
         \begin{array}{rl}
         0, & (i,j) \in r_s \\
         1,    & otherwise,   \\
         \end{array}
   \right.
   \label{eq:dropblock mask}
\end{equation}
and obtain dropped feature maps $\boldsymbol{D}_n$ by applying the mask on the low-level features $\boldsymbol{B}_n$ with normalization:
\begin{equation}
   \boldsymbol{D}_n = \boldsymbol{B}_n \times M \times {\rm Count(M)} / {\rm Count\_ones(M)} ,
   \label{eq:dropblock feature}
\end{equation}
where ${\rm Count}(\cdot)$ and ${\rm Count\_ones}(\cdot)$ denote the number of all elements and the number of elements with value one, respectively. 

Different from the original random dropblock, our ROI guided dropblock can directly erase the informative part and encourage the network to find more discriminative regions, which achieves higher accuracy. Table \ref{tab:ablation study on refinement} shows the comparison of the random dropblock and our ROI guided dropblock. Note that we only conduct dropblock refinement in the training process, but in the test process we skip this operation.

\begin{table*}[htbp]
  \begin{center}
  \small
    \begin{tabular}{|c|c|c|c|c|c|c|}
    \hline
    \multirow{2}*{Method} & \multirow{2}*{Base} & \multirow{2}*{Pre-trained} &  \multirow{2}*{Image resolution} & \multicolumn{3}{c|}{Accuracy (\%)} \\
    \cline{5-7}       
     &  &  &  &  CUB-200-2011 & Stanford Cars & FGVC-Aircraft \\
    \hline
    \hline
    FT VGGNet~\cite{wang2018learning} & VGG19 & ImageNet & $448\times448$ & 77.8 & 84.9 & 84.8 \\
    FT ResNet~\cite{simonyan2014very} & ResNet50 & ImageNet & $448\times448$ & 84.1 & 91.7 & 88.5 \\
    B-CNN~\cite{lin2015bilinear} & VGG16 & ImageNet & $448\times448$ & 84.1 & 91.3 & 84.1 \\
    MA-CNN ~\cite{zheng2017learning} & VGG19 & ImageNet & $448\times448$ & 86.5 & 92.8 & 89.9 \\
    DFL~\cite{wang2018learning} & ResNet50 & ImageNet & $448\times448$ & 87.4 & 93.1 & 91.7 \\
    NTS~\cite{yang2018learning} & ResNet50 & ImageNet & $448\times448$ & 87.5 & 93.9 & 91.4 \\
    DCL~\cite{chen2019destruction} & ResNet50 & ImageNet & $448\times448$ & 87.8 & 94.5 & \underline{93.0} \\
    TASN~\cite{zheng2019looking} & ResNet50 & ImageNet & $448\times448$ & 87.9 & 93.8 & - \\
    WS-CPM~\cite{ge2019weakly} & GoogleNet & ImageNet+COCO & Shorter side is 800 & \textbf{90.3} & - & - \\
    \hline
    AP-CNN (one stage) & VGG19 & ImageNet & $448\times448$ & 85.4 & 93.2 & 91.5 \\
    AP-CNN (two stages) & VGG19 & ImageNet & $448\times448$ & 86.7 & \underline{94.6} & 92.9 \\
    AP-CNN (one stage) & ResNet50 & ImageNet & $448\times448$ & 87.2 & 93.6 & 92.2 \\
    AP-CNN (two stages) & ResNet50 & ImageNet & $448\times448$ & \underline{88.4} & \textbf{95.4} & \textbf{94.1} \\
    \hline
    \end{tabular}%
  \end{center}
  \caption{Comparison results on CUB-200-2011, Stanford Cars, and FGVC-Aircraft datasets. The best and second-best results are respectively marked in bold and underlined fonts.}
  \label{tab:main}%
\end{table*}%

\textbf{ROI Guided Zoom-in.} We merge ROIs from all pyramid levels to learn the minimum bounding rectangle of the input image in a weakly-supervised way,
and get $[t_{x1}, t_{x2}, t_{y1}, t_{y2}]$ denoting the minimum and maximum coordinates in terms of $x$ and $y$ axis of the merged bounding rectangle, respectively. Then we extract this region out from the dropped feature maps $\boldsymbol{D}_n$, and enlarge it to the same size as $\boldsymbol{D}_n$ to get the zoom-in features $\boldsymbol{Z}_n$:
\begin{equation}
   \boldsymbol{Z}_n = \varphi (\boldsymbol{D}_n[t_{y1}:t_{y2}, t_{x1}:t_{x2}]),
   \label{eq:crop_resize}
\end{equation}
where $\varphi$ represents the bilinear upsample operation. The refined features $\boldsymbol{Z}_n$ is sent to the refined-stage to conduct further prediction. The final prediction is made by averaging the raw-stage prediction and refined-stage prediction.



\section{Experimental Results and Discussions}\label{experiments}
We conduct experiments on three FGVC benchmark datasets, including CUB-200-2011, Stanford-Cars, and FGVC-Aircraft. All the datasets contain a set of sub-categories of the same super-category. The following is a brief description of these datasets:

\textbf{CUB-200-2011 }\cite{wah2011caltech} has 11,778 images from 200 classes officially split into 5,994 training and 5,794 test images.

\textbf{Stanford-Cars }\cite{krause20133d} has 16,185 images from 196 classes officially split into 8,144 training and 8,041 test images.

\textbf{FGVC-Aircraft }\cite{maji2013fine} has 10,000 images from 100 classes officially split into 6,667 training and 3,333 test images.

\subsection{Implement Details}\label{experiment:implement}
We implement AP-CNN on 50-layer ResNet~\cite{he2016deep} pre-trained on ImageNet. Specifically, we choose the last output feature of the residual block conv3, conv4 and conv5 in ResNet50 to establish pyramidal hierarchy, denoting as $B_3, B_4, B_5$ respectively. We do not include conv1 and conv2 into the pyramid because of their large memory footprint. The refinement operation is conducted on $B_3$ (for detailed information of this choice, please refer to Table \ref{tab:refinement on imagefeature}). The input images are resized into $448\times 448$, which is standard in the literature. We do not use extra bounding box/part annotations and compare our method with other weakly supervised approaches. 
We respectively assign anchors with single scales of {64, 128, 256} and 1:1 ratio for each pyramidal level and choose the top 5, 3, 1 anchors with the highest activation value as potential refinement candidates. The IOU threshold in NMS operation is set as 0.05 and the dropblock rate in Algorithm~\ref{alg:refinement} is set as $\{30\%, 30\%, 0\%\}$. Note that most of the hyperparameters of AP-CNN are involved in the process of getting anchors and the dropblock operation, we set them with empirical experiences.

We use open-sourced Pytorch as our code-base, and train all the models on a single GTX 1080Ti GPU. Optimization is performed using Stochastic Gradient Descent with momentum 0.9 and a minibatch size of 16. The initial learning rate is set to 0.001 and drops to 0 using cosine anneal schedule. All models are trained for 100 epochs.

\subsection{Comparison with State-of-the-Art Methods}\label{experiment:comparison}
Table~\ref{tab:main} lists the performance evaluations on three aforementioned benchmark datasets. Each column includes 7 to 9 representative weakly supervised methods that have reported evaluation results on the corresponding datasets, including fine-tuned baselines, feature encoding methods and region locating methods. We display the results of our model based on the VGG16 and ResNet50 backbone.
Compared with the above FGVC works, our AP-CNN achieves significant performance improvement on all the three datasets. The evaluation results can be summarized as follows:
\begin{itemize}
   \item On the CUB-200-2011 dataset, our AP-CNN (one stage) achieves a significant improvement from the corresponding backbones, with clear margins of 7.6\% and 3.1\% on VGG19 and ResNet50, respectively. By applying ROI guided refinement, AP-CNN (two stages) reaches 88.4\% accuracy on ResNet50, which outperforms the existing methods using the same backbone, pre-trained datasets, and image resolution. Note that the WS-CPM model currently gets the highest classification accuracy with 90.3\%, which mainly benefits from the extra pre-trained data and the high input resolution.
   \item On the Stanford Cars dataset, currently the state-of-the-art accuracy is achieved by the DCL model with 94.5\%. Our method outperforms DCL for a clear margin (0.9\% relative gain) with accuracy 95.4\%. 
   \item On the FGVC-Aircraft dataset, our method again reaches the best accuracy of 94.1\%. Compared with the leading result achieved by DCL, the relative accuracy gain is 1.1\%, which confirms the significance of our method.
\end{itemize}
Overall, the proposed AP-CNN benefits from two aspects: 1) By establishing the pyramidal hierarchy on CNN backbones, we extract multi-scale features guided by individual attention activations, which can distill both the high-level semantic and the low-level detailed information for better classification and precise localization. 2) Conducting the ROI guided refinement (aligning features with background noise excluded by ROI guided zoom-in, and enhancing discriminative parts on features by ROI guided dropblock) on the refined-stage can also contribute to performance improvement.

\subsection{Ablation Studies}\label{experiment:ablation}
We conduct ablation studies to analyze the contribution of each component. The following experiments are all conducted on the CUB-200-2011 dataset and we use ResNet50 as the backbone if not particularly mentioned. 

\textbf{Effect of the Pyramidal Hierarchy.}
We investigate the effect of constructing the pyramidal hierarchy on CNN backbones by comparing the performances obtained by the backbone, by the feature pyramid structure (FP), and by the attention pyramid structure (AP) on the VGG and ResNet network. As shown in Table~\ref{tab:pyramidal hierarchy}, FP leads significant performance improvement compared to the baseline, and AP further raises the accuracy by enhancing the correlations between features. The results confirm that the pyramidal architecture with multi-level information is essential in the FGVC task. 

\textbf{Arrangement of the Attention Components. }
The attention union in AP-CNN consists of two parts: the spatial attention pyramid and the channel attention pathway. We conduct experiments to demonstrate their effectiveness by evaluating the classification accuracy, the mean Intersection-over-Union (mIoU) and the recall rate with ground-truth bounding boxes. Table~\ref{tab:attention} shows that using the spatial attention set or the channel attention set alone or their combination can only contribute limited improvement in classification accuracy. We add the additional bottom-up pathways to enhance the correlationship between the neighboring pyramidal levels, which can be constructed by either channel attetion or spatial attention. The spatial attention based bottom-up pathway will make the activation maps become similar, which is not consistent to our motivation and therefore yields poor mIoU and recall rate. The channel attention based bottom-up pathway is reasonable, as the FPN has made the features from different levels aligned in channels. The experimental results confirm that the channel attention pathway is an appropriate choice.

\begin{table}[t]
  \begin{center}
  \small
    \begin{tabular}{|l|c|c|}
    \hline
    Method & Base Model & Accuracy (\%)   \\
    \hline
    \hline
    Baseline & VGG19 / ResNet50 &   77.8 / 84.1 \\
    FP       & VGG19 / ResNet50 &   83.3 / 86.6 \\
    FP + AP  & VGG19 / ResNet50 &   \textbf{85.4 / 87.2} \\
    \hline
    \end{tabular}%
  \end{center}
  \caption{Comparison results on backbones with/without pyramidal hierarchy structure. FP: Feature pyramid. AP: Attention pyramid.}
  \label{tab:pyramidal hierarchy}%
\end{table}%

\begin{table}[t]
  \begin{center}
  \small
    \begin{tabular}{|l|c|c|c|}
    \hline
    Method                       & Accuracy (\%)   & mIoU (\%)    & Recall (\%) \\
    \hline
    \hline
    FPN                       &        86.6  &     -   &    -        \\
    FPN + C                &        86.8  &     -   &    -        \\
    FPN + S                &        86.7  &     54.9 &   73.6        \\
    FPN + S + C         &        86.7  &     54.9 & \textbf{74.5}    \\
    \hline
    FPN + C + SP         &        86.9  &    54.1 & 72.0    \\
    FPN + S + CP  & \textbf{87.2}   & \textbf{56.4} &    74.0     \\
    \hline
    \end{tabular}%
  \end{center}
  \caption{Comparison results on different ways of getting Attention. C: Channel attention. S: Spatial attention. SP: Bottom-up spatial attention pathway. CP: Bottom-up channel attention pathway.}
  \label{tab:attention}%
\end{table}%

\begin{table}[t]
  \begin{center}
  \small
    \begin{tabular}{|c|c|c|c|}
    \hline
    Erasing       & Zoom-in      &   Guidance      & Accuracy (\%)   \\
    \hline
    \hline
    $-$           & $-$          &    ROI / Random    & 87.2 / 87.2 \\
    \checkmark       & $-$          &    ROI / Random    & 87.4 / 87.1  \\
    $-$           & \checkmark    &    ROI / Random   & 87.6 / 87.3  \\
    \checkmark       & \checkmark    &    ROI / Random   & \textbf{88.4} / 87.6  \\
    \hline
    \end{tabular}%
  \end{center}
  \caption{Contribution of each refinement component.}
  \label{tab:ablation study on refinement}%
\end{table}%

\begin{figure*}[!t]
\begin{center}
   \includegraphics[width=1.0\linewidth]{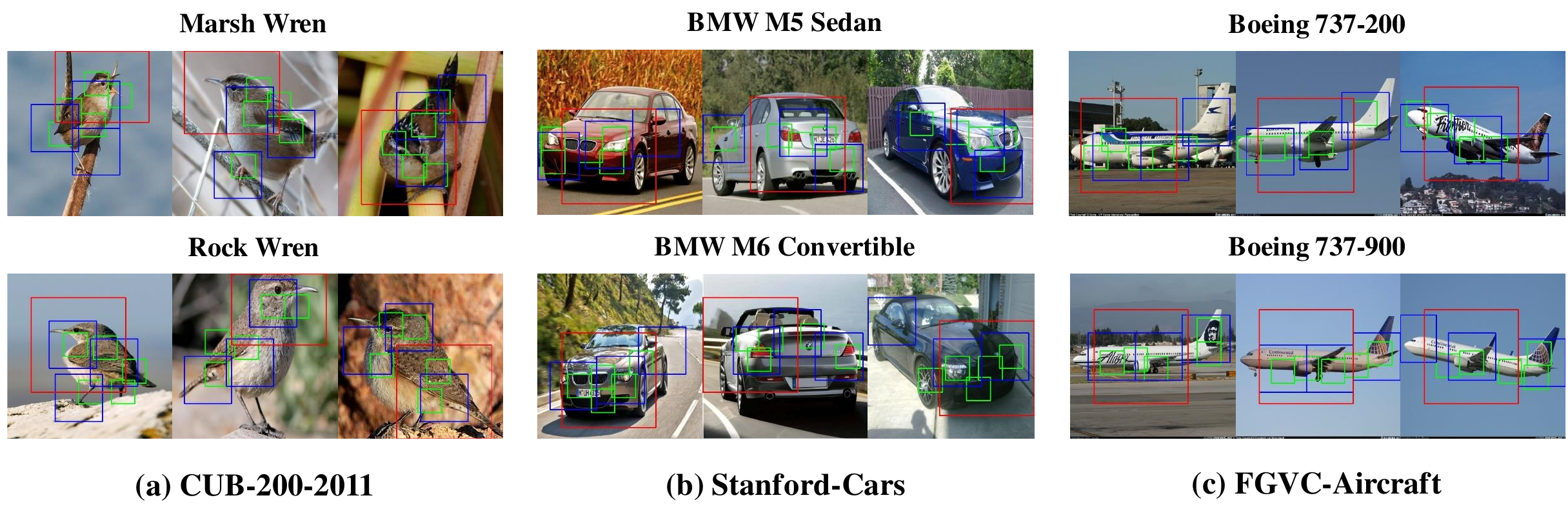}
\end{center}
\vspace{-3mm}
   \caption{Visualization of ROI pyramid from two similar sub-categories on (a) CUB-200-2011, (b) Stanford-Cars, and (c) FGVC-Aircraft dataset. Rectangles with different colors are from distinctive levels in the pyramidal hierarchy.}
\label{fig:fig4}
\vspace{-2mm}
\end{figure*}

\textbf{Contribution of the Refinement Components. }
As described above, our ROI guided refinement mainly consists of two operations, including the ROI guided dropblock and the ROI guided zoom-in. We conduct the ablation studies upon them, and compare the ROI guided methods with the corresponding random operations. As shown in Table~\ref{tab:ablation study on refinement}, ROI guided methods have advantages compared with the random ones in many aspects, and both the two refine components are effective in our refinement process.

\textbf{Model Complexity. }
Our refinement operations can be theoretically conducted on any low-level position of the network (\eg, the input and the feature maps from conv1, conv2) by sampling the ROIs into different scales. Table \ref{tab:refinement on imagefeature} compares the classification accuracies and the training time costs (on a single GTX 1080Ti GPU) among different refinement positions. We consequently refine the low-level features from Conv3 with the consideration of both accuracy and efficiency.

\begin{table}[t]
  \begin{center}
  \small
    \begin{tabular}{|c|c|c|}
    \hline
    Refinement position & Accuracy (\%) & Time cost \\
    \hline
    \hline
    Input image   &    88.1          & 527s / epoch \\
    Conv1 feature &    87.9          & 525s / epoch \\
    Conv2 feature &    88.1          & 458s / epoch \\
    Conv3 feature &   \textbf{88.4} & \textbf{389s / epoch} \\
    \hline
    \end{tabular}%
    \end{center}
  \caption{Comparison between refinement on input image and low-level features.}
  \label{tab:refinement on imagefeature}%
\end{table}%

\begin{table}[t]
  \begin{center}
  \small
    \begin{tabular}{|c|c|c|c|c|c|}
    \hline
    \multirow{2}*{Method} & \multirow{2}*{GFlops} & \multirow{2}*{Params} & \multicolumn{3}{c|}{Accuracy (\%)}  \\
    \cline{4-6}   
     & & & Birds  &  Cars  & Airs  \\
    \hline
    \hline
    Baseline & 16.48        & 25.56M    & 84.1         &   91.7       & 88.5 \\
    A        & 19.64        & 27.96M    & 86.7         &   93.1       & 91.5 \\
    B        & 19.64        & 27.96M    & 87.2         &   93.6       & 92.2 \\
    C        & 31.93        & 27.96M    &\textbf{88.4} &\textbf{95.4} &\textbf{94.1}\\
    \hline
    \end{tabular}%
  \end{center}
  \caption{Detailed information of the most contributed parts. A: AP-CNN with two most contributed parts (the bottom-up pathway and the ROI guided refinement) removed, B: A + bottom-up pathway, C: B + ROI guided refinement.}
  \label{tab:info of contribute parts}%
\vspace{-2mm}
\end{table}%

In summary, the efficiency of the proposed method can benefit from three aspects: a) the bottom-up pathway only sums two channel attention masks without additional parameters, b) the raw-stage and the refined-stage are based on the same network with shared parameters, and c) the refinement is conducted on low-level features rather than inputs.
Table \ref{tab:info of contribute parts} shows consistent improvements yielded by the most contributed parts (the bottom-up pathway and the ROI guided refinement) on the aformentioned FGVC datasets, with limited increase of the amount of parameters and computational cost.

\subsection{Visualization}\label{experiment:visualization}
Figure~\ref{fig:fig4} visualizes the ROI pyramid learned by the AP-CNN. In each line, we randomly select three test images from one specific ``wren'' species from the CUB-200-2011 dataset, one ``BMW'' series from the Stanford-Cars dataset, and one ``Boeing'' series from the FGVC-Aircraft dataset. We use the red, blue and green boxes to denote the most activated regions, with the red ones representing the high-level ROIs with big anchor size, to the green ones representing the low-level ROIs with small anchor size. It can be intuitively observed that the localized regions are indeed informative for fine-grained classification, and the ROIs from different pyramidal levels can focus on more distinctive parts due to their particular receptive field.

\section{Conclusion}\label{conclution}
In this paper, we propose an attention pyramid network for fine-grained image classification without extra annotations. This is conducted via building a pyramidal hierarchy upon CNN which consists of a top-down feature pathway and a bottom-up attention pathway to deliver both the high-level semantic and the low-level detailed information. ROI guided refinement, which enhances the discriminative local activations and aligns the features with the background noises eliminated, is conducted to further improve the performance. Experiments on CUB-Bird, Stanford-Cars, and FGVC-Aircraft demonstrate the superiority of our method.

{\small
\bibliographystyle{ieee_fullname}
\bibliography{mybibfile}
}

\end{document}